\documentclass[runningheads]{llncs}
\usepackage{amsfonts}
\usepackage{graphicx}
\usepackage{amsmath,bm}
\usepackage{hyperref}
\usepackage{multirow}
\usepackage{subfigure}
\usepackage{float}
\usepackage[noend]{algpseudocode}
\usepackage{algorithmicx,algorithm}

\begin{document}
\title{Scene Text Recognition with Image-Text Matching-guided Dictionary 
}
%
%
\author{
Jiajun Wei\inst{1} \and
Hongjian Zhan\inst{1, 2} \and
Xiao Tu\inst{1} \and
Yue Lu\inst{1\thanks{Corresponding author}} \and
Umapada Pal\inst{3}
}
%

\institute{
Shanghai Key Laboratory of Multidimensional Information Processing, East China
Normal University, Shanghai, China\\
\and
Chongqing Institute of East China Normal University. Chongqing. 401120. China
\and
CVPR Unit, Indian Statistical Institute, Kolkata, India\\
\email{
jjwei@stu.ecnu.edu.cn, ecnuhjzhan@foxmail.com, xtu@cee.ecnu.edu.cn, ylu@cs.ecnu.edu.cn, umapada@isical.ac.in
}
}



%
\maketitle              
\begin{abstract}

Employing a dictionary can efficiently rectify the deviation between the visual prediction and the ground truth in scene text recognition methods. However, the independence of the dictionary on the visual features may lead to incorrect rectification of accurate visual predictions. In this paper, we propose a new dictionary language model leveraging the \textbf{S}cene \textbf{I}mage-\textbf{T}ext \textbf{M}atching(SITM) network, which avoids the drawbacks of the explicit dictionary language model: 1) the independence of the visual features; 2) noisy choice in candidates etc. The SITM network accomplishes this by using Image-Text Contrastive (ITC) Learning to match an image with its corresponding text among candidates in the inference stage. ITC is widely used in vision-language learning to pull the positive image-text pair closer in feature space. Inspired by ITC, the SITM network combines the visual features and the text features of all candidates to identify the candidate with the minimum distance in the feature space. Our lexicon method achieves better results(93.8\% accuracy) than the ordinary method results(92.1\% accuracy) on six mainstream benchmarks. Additionally, we integrate our method with ABINet and establish new state-of-the-art results on several benchmarks.


\keywords{Dictionary Language Model \and Scene Image-Text Matching \and Image-Text Contrastive Learning \and Scene Text Recognition.}
\end{abstract}
%
%
%
\section{Introduction}
Deep learning-based scene text recognition has been developed for years. The accuracy of scene text recognition has vastly increased as the appropriate design of model architecture and the expansion of model size. Previous methods\cite{ref_lncs18,ref_lncs11,ref_lncs12,ref_lncs14} can address a variety of recognition issues, but the inherent ambiguities, such as complicated background or diversity of font, etc, render the recognized results inaccurate. 

Due to the unique characteristics of text recognition, it is feasible to employ human language priors to rectify the output of a vision recognition model. Utilizing a pre-trained language model is one of the common methods. Fang et al\cite{ref_lncs44} pre-train a language model using WikiText-103\cite{ref_lncs45}. The pre-trained language model rectifies the visual prediction through learning grammar and the construction of words in the human language system. Another popular approach is to search for a word that has minimum edit distance(Levenshtein distance\cite{ref_lncs46}) with the visual prediction in a dictionary. Nguyen et al\cite{ref_lncs48} present a method for incorporating a dictionary into the training pipeline. They use the dictionary to generate a certain number of candidates and then output the most compatible one with the highest compatibility scores in a probability matrix \textbf{P}, which is generated by the visual feature $\bm F_v$. But they still disregard the interaction between visual features and text features in the inference stage.\\
\indent The aforementioned methods utilizing explicit language models have several problems. First, regardless of the pre-trained language model or dictionary language model, the independence of the language model from the visual feature may erroneously rectify the correct prediction results. Second, it is illogical to utilize human language priors to rectify texts that appear infrequently or have no linguistic information($e.g.$ ngee, tsc), since neither a pre-trained language model nor a dictionary can rectify texts without human language logic. 
\begin{figure}[]
\centering
\includegraphics[width=1\linewidth]{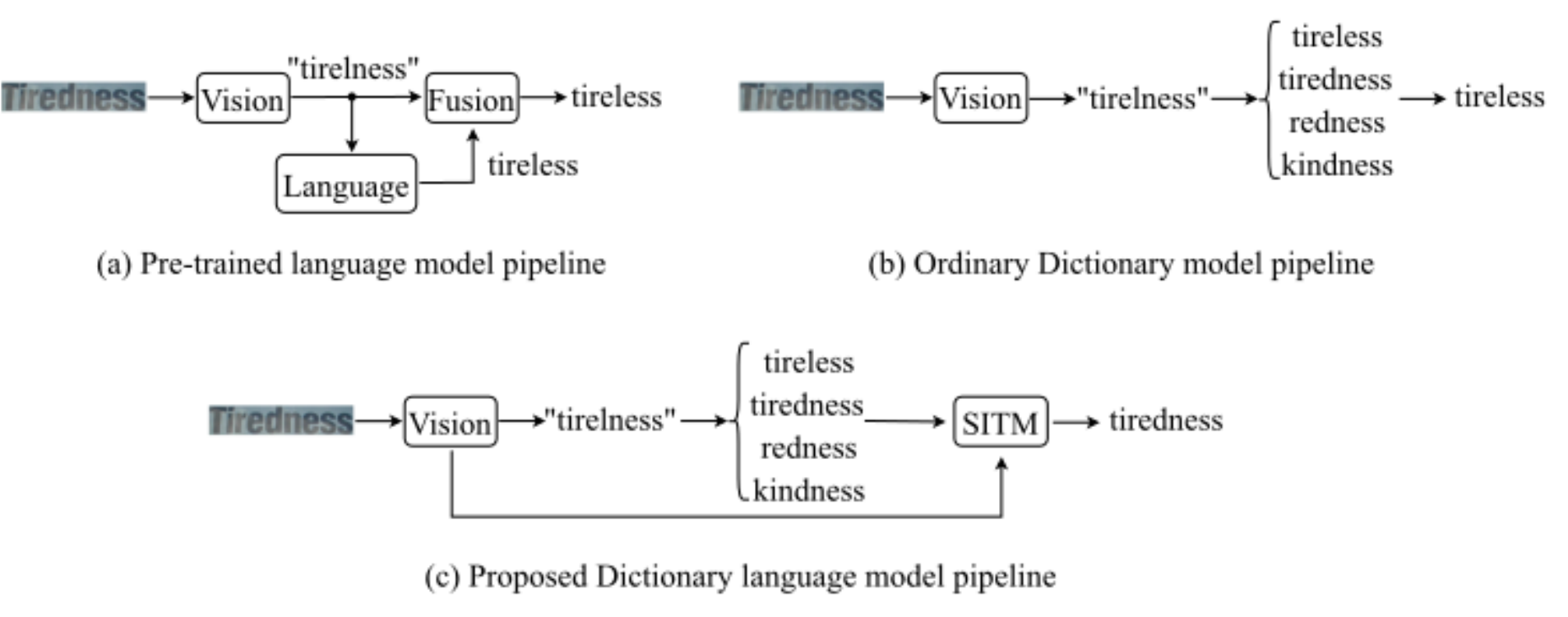}
\caption{Comparison with different pipelines.}
\label{intro_small}
\end{figure}

In this paper, we present an effective method for incorporating a dictionary into scene text recognition that possesses two advantages: 1) taking visual prediction into account. When generating candidates for the visual prediction, the visual prediction is also included in the candidate set. In this case, the initial visual prediction has a probability to be the ultimate outcome. 2) Integrating visual features
into the inference stage. Image-Text Contrastive(ITC) Learning is an unsupervised learning method that aims to make positive image-text pairs have higher similarity scores. Inspired by ITC learning, we additionally integrate a \textbf{S}cene \textbf{I}mage-\textbf{T}ext \textbf{M}atching(SITM) network to match visual features and text features by ITC Learning. Nevertheless, when we merely employ other label texts in the same batch as negatives, the image-text matching accuracy in the inference stage is not as good as the training stage. We address the problem by generating hard negatives that resemble the shape of label texts. The difference between three methods is depicted in Fig. \ref{intro_small}\\
\indent The main contributions of this paper are summarized as follows: 1) we propose a novel method to integrate a dictionary into scene text recognition that avoids the drawbacks of an ordinary dictionary language model. 2) We also offer a new strategy that employs labels to generate resemblant words as hard negatives in the SITM training stage. 3) A Scene Image-Text Matching Module is introduced, which matches positive image-text pairs in the inference stage. 

\section{Related Work}
\subsection{Scene Text Recognition}
\subsubsection{Language-free Methods.}
Language-free approaches typically provide a prediction based on visual features, regardless of context information. CTC-based methods\cite{ref_lncs21} 
utilize CNN to extract visual features, RNN to model sequence features, and CTC loss to train the entire recognition network end-to-end \cite{ref_lncs23,ref_lncs24,ref_lncs25}. Segmentation-based methods\cite{ref_lncs22} segment each character region before classifying and recognizing. The recognition results of all character areas compose the entire text sequence\cite{ref_lncs26,ref_lncs27,ref_lncs28}. However, due to the absence of context information interaction, these approaches cannot attain exceptional performance.

\subsubsection{Language-based Methods.}

In previous works, \cite{ref_lncs2,ref_lncs3} use explicit language models to improve model recognition accuracy. CNN is employed to extract visual features to predict bags of N-grams of text strings. Recently, \cite{ref_lncs44} regards the explicit language model as a spell checker to rectify visual prediction results. Some implicit language-based approaches connect visual features with context information by utilizing RNN\cite{ref_lncs7,ref_lncs8} or attention mechanisms \cite{ref_lncs9,ref_lncs10}. First, an image encoder is employed to extract features from word images, follower by an attention-based method for integrating visual features and context information. \cite{ref_lncs11,ref_lncs12,ref_lncs13,ref_lncs18} focus on relevant information from 1D image features, and \cite{ref_lncs14,ref_lncs15,ref_lncs16,ref_lncs17} from 2D image features. Some performance-enhancing approaches focus on learning new feature representations. \cite{ref_lncs4,ref_lncs5,ref_lncs6} train their models by sequence contrastive learning, masked image modeling, and a mix of the two, respectively. 

\subsection{Vision-Language Learning}
There are two categories of visual-language representation learning. In the first category, text features and image features are fused using a multi-mode encoder \cite{ref_lncs35,ref_lncs36,ref_lncs37,ref_lncs38}. This type of approaches has achieved outperformance in downstream tasks such as NLVR\cite{ref_lncs39} and VQA\cite{ref_lncs40}. The Second category focus on learning separate texts and images encoders\cite{ref_lncs41,ref_lncs42}. CLIP\cite{ref_lncs43} employs contrastive loss to train the image encoder and text encoder on a massive quantity of network image-text pairs. We opt for the second category to reuse the visual encoder trained in the recognition stage instead of training a new visual encoder. Then, the text encoder is trained from scratch in the matching stage. 


\section{Method}

We propose a new method to incorporate a dictionary into scene text recognition. The dictionary is used to generate the certain number of candidates, which will subsequently be matched with the visual features by SITM to output the candidate with the highest similarity score. In this section, the details of the overall architecture are presented. We will also describe the Resemblant word generation strategy and the SITM network. The objective training function is finally introduced.

\begin{figure}[H]
\centering
\includegraphics[width=1\linewidth]{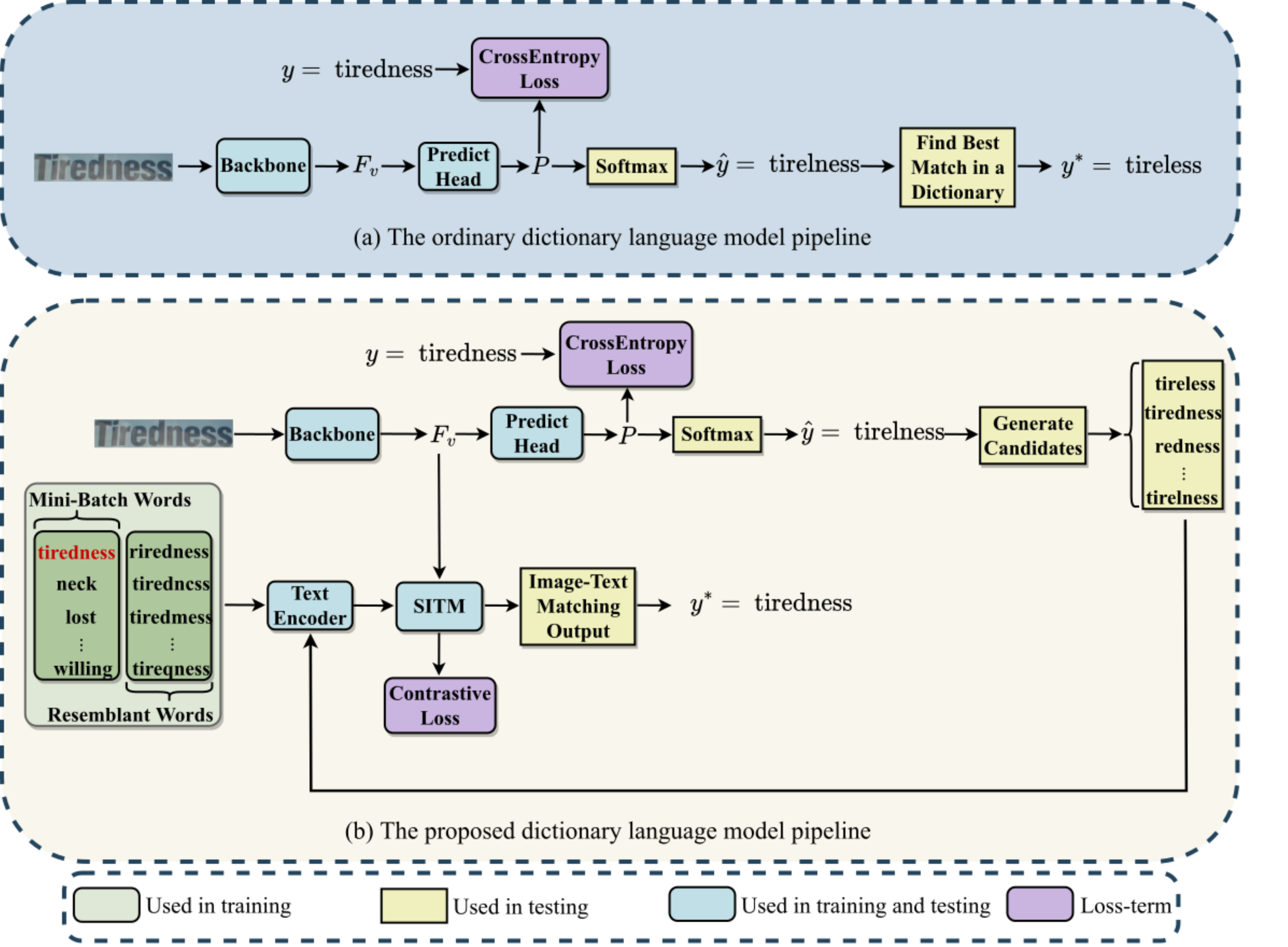}
\caption{Ordinary dictionary language model pipeline (a) and proposed dictionary language model pipeline (b). In the ordinary pipeline, the prediction is forced to be one with the smallest edit distance in the dictionary. In the proposed pipeline, the ultimate prediction is determined by SITM}
\label{over_arch}
\end{figure}

\subsection{Overall Architecture}

\begin{figure}[]
\centering
\includegraphics[width=1\linewidth]{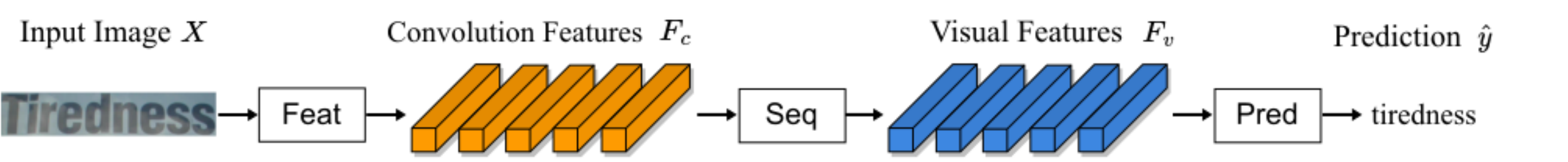}
\caption{General Vision Architecture of Scene Text Recognition.}
\label{hard_dict_negatives}
\label{vision_pipeline}
\end{figure}

As can be seen in Fig. \ref{vision_pipeline}, a general scene text recognition framework usually consists of a feature extraction module, a sequence modeling module, and a prediction module, which was proposed by Baek et al\cite{baek2019wrong}. Our proposed dictionary method can combine any scene text recognition method with the above framework. We utilize the output of the sequence modeling as the visual features $\bm F_{v} $. Specifically in this paper, we employ the vision module of the ABINet\cite{ref_lncs44} as our baseline network.

Fig. \ref{over_arch}b describes the procedure of our proposed recognition pipeline, which employs a forward-forward method. For the initial forward, we input the image, generate visual prediction $\hat{y}$ with the text recognition network, and then utilize $\hat{y}$ to find candidates with the top N smallest edit distance in the dictionary. The candidate set is comprised of the N candidates and the visual prediction $\hat{y}$. For example, if $\hat{y}= tirelness$, the candidates will be: \emph{tireless, tiredness, redness, ..., kindness, tirelness}. For the second forward pass, the inputs are the image and the candidates obtained from the first forward. Then, utilizing the SITM network to match the text in the candidates with the image, and output the text with the highest similarity. The inference procedure is depicted in Algorithm 1.

\begin{algorithm}[]
\caption{Inference procedure} 
\hspace*{0.02in} {\bf Input:} 
 $x$: Input Image; $n$: Forward State \\
\hspace*{0.02in} {\bf Output:} 
Prediction $y^{*}$
\begin{algorithmic}[1]
\State initial $n = 1$ 
\For{$n = 1, 2$} %
    \If{$n = 1$} %
       \State Input $x$ to get $\hat{y} = {\bf V}(x)$, where $\bf{V}$ is the vision module
       \State Construct candidate set $C$ of $\hat{y}$ using dictionary
    \Else
       \State Input $x$ and $C$ to calculate the similarity scores $S$
       \State Get the $c \in C$ with the highest score in $S$ as the final prediction $y^{*}$
    \EndIf
\EndFor
\State \Return $y^{*}$
\end{algorithmic}
\end{algorithm}

During training, we generate text candidates for Image-Text Contrastive Learning by using labels in the same mini-batch. In addition, we create a certain number of resemblant words as hard negatives. A contrastive loss function, which is defined based on the similarity cross-entropy function depicted in Section 3.3, and the recognition loss are then employed for training.


\subsection{Resemblant Words Generation}

\begin{figure}[]
\centering
\includegraphics[width=1\linewidth]{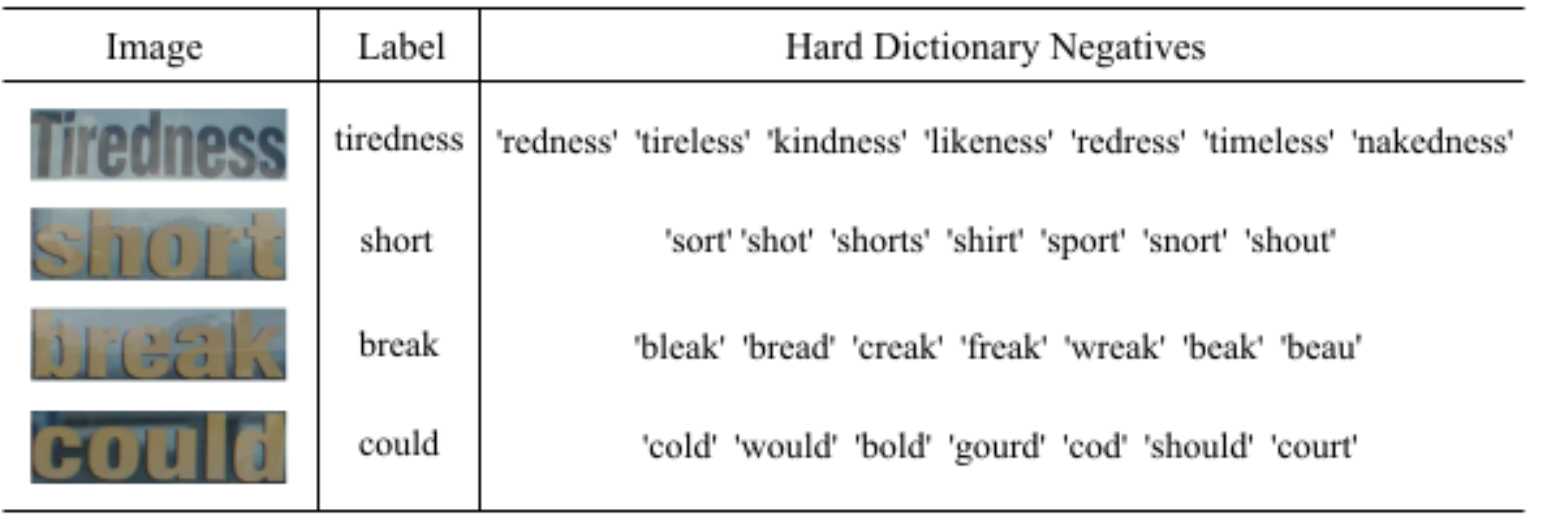}
\caption{Qualitative hard negatives in the inference stage.}
\label{hard_dict_negatives}
\end{figure}

There is a gap between the SITM training stage and the inference stage if we utilize the normal contrastive learning method. Specifically, in the training stage, the negatives are the other labels in the same mini-batch for a single image. For example, if a mini-batch contains: \emph{tiredness, kills, short, break, could, save, \textbf{your}, life}, the text negatives are \emph{tiredness, kills, short, break, could, save, life} and the text positive is \emph{\textbf{your}} for the image \emph{\textbf{your}}. However, in the inference stage, the negatives are candidates from the dictionary with the top N smallest edit distance, which is similar to the ground truth. For example, for the image \emph{\textbf{your}}, the negatives in the inference stage are \emph{ pour, you, tour, hour, dour, sour, four}. Fig. \ref{hard_dict_negatives} exhibits some hard negatives for the labels in the test set. We find the gap would cause some mismatches between image-text pairs in the inference stage and degrade the performance of the dictionary. 

We address the problem with our proposed Resemblant Words Generation strategy. When we train the SITM network, in addition to using the text labels corresponding to other images in the same batch as negatives, we present a strategy for constructing hard negatives using labels. Specifically, we initially establish a similar character lookup table containing five similar characters for each English character. We observe the difference between visual prediction and ground truth and record the wrong predicted characters as the composition of the lookup table. For character $a$, we select $d, e, o, q$ and $u$ as the similar characters. Then we randomly replace a character in a label having a similar appearance. For example, if $\rm y$ = $tiredness$ and the number of the resemblant words is $4$, the hard negatives will be $riredness$, $tiredncss$, $tiredmess$, $tireqness$. 


\subsection{Scene Image-Text Matching Module}

\begin{figure}[]
\centering
\includegraphics[width=1\linewidth]{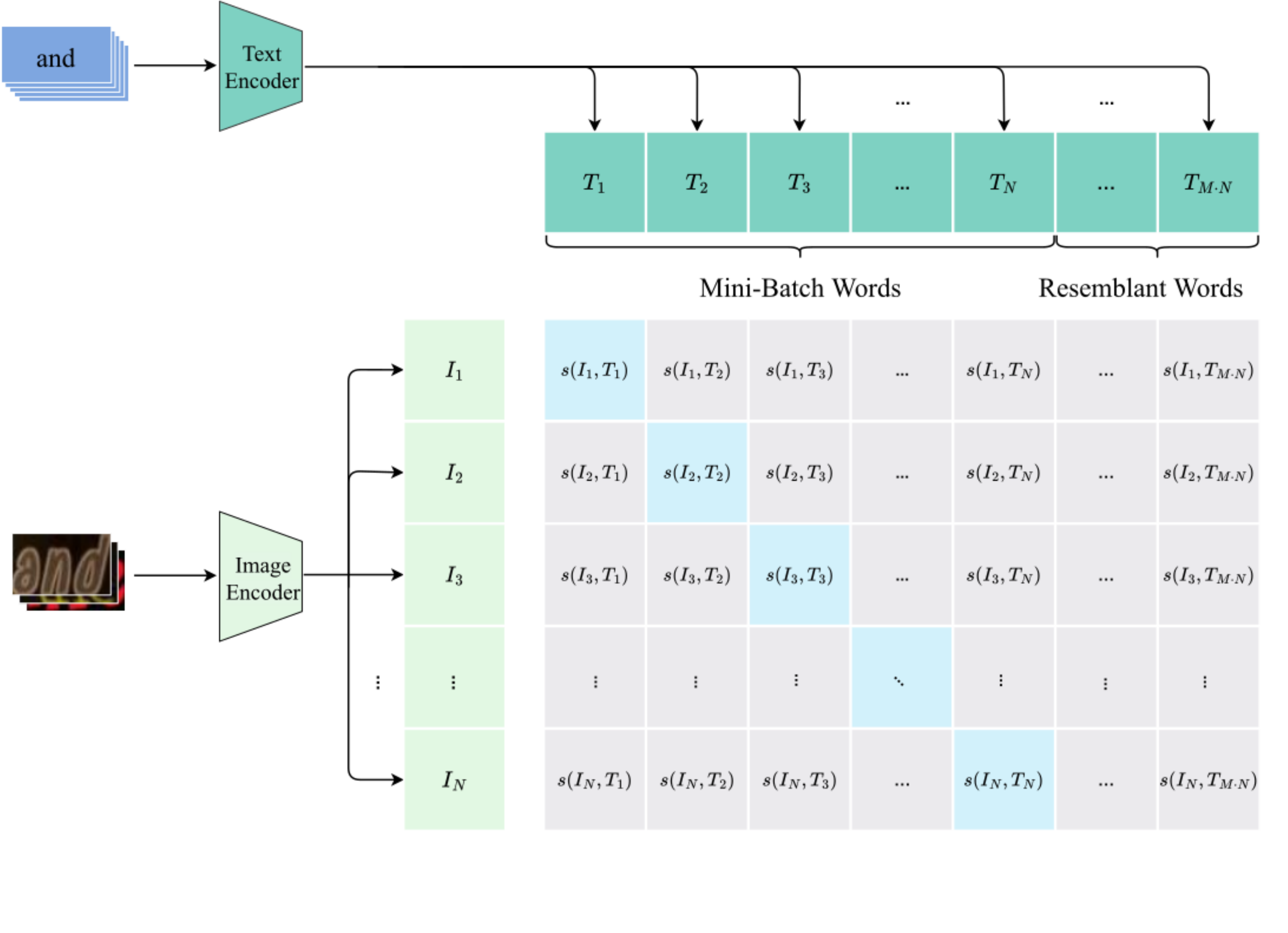}
\caption{Scene Image-Text Matching Module Architecture.}
\label{scene_itm_arch}
\end{figure}


The Scene Image-Text Matching module contains image encoder, text encoder and Scene Image-Text Contrastive Learning module. The image encoder consists of a backbone network that shares parameters with the backbone network of the recognition module and a parallel attention layer that is used to convert visual features to sequence features. The text encoder consists of two layers of transformer encoder. Scene Image-Text Contrastive Learning module consists of two liner layers. Image sequence features $\bm{I}\in\mathbb{R}^{L\times C}$ and text sequence features $\bm{T} \in\mathbb{R}^{L\times C}$ are obtained by image encoder and text encoder, respectively. The image features $\bm I$ and text features $\bm T$ pass through the linear projection layer before Image-Text Contrastive Learning.
Fig. \ref{scene_itm_arch} shows the details of our SITM module.

\begin{figure}[H]
\centering
\includegraphics[width=0.8\linewidth]{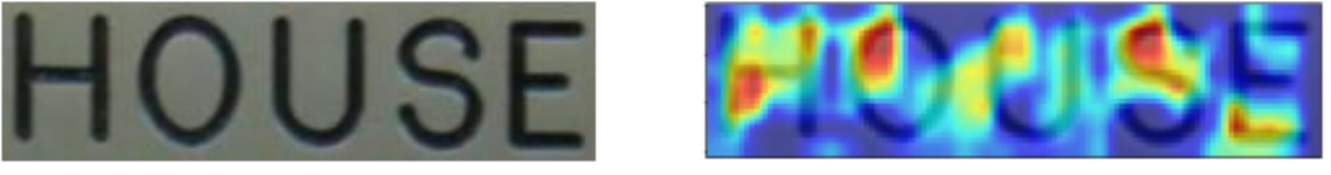}
\caption{An example of the parallel attention layer Gradient-weighted Class Activation Mapping. The left image is the input and the right image is the activation mapping.}
\label{sitm_cam}
\end{figure}

During the training stage, we employ the image-text contrastive learning task in vision-language learning to complete the scene image-text matching task. Image-Text Contrastive Learning aims to learn a representation of distinct modal features. It learns the cosine similarity function $ s({\bm I}, {\bm T})=\frac{l_v({\bm I}) \cdot l_t({\bm T})}{||l_v({\bm I})|| \cdot ||l_t({\bm T})||}$, where $l$ represents linear layer and $\bm{I}\in\mathbb{R}^{L \times C}$, $\bm{T}\in\mathbb{R}^{L \times C}$ represent image features and text features, respectively. The matched image-text pair will have a higher similarity score. We calculate image-to-text(i2t) and text-to-image(t2i) similarity and normalize the results using softmax. The formulas are as below:
\begin{equation}
    p^{i2t}_m(I) = \frac{{\rm exp}(s(I, T_m) / \tau)}{\sum_{n=1}^{MN} {\rm exp}(s(I, T_n) / \tau},
    \quad
    p^{t2i}_m(T) = \frac{ {\rm exp}(s(T, I_m) / \tau)}{\sum_{n=1}^{N} {\rm exp}(s(T, I_n) / \tau)},
\end{equation}
where $\tau$ is a temperature parameter, $m$ is the order indication of the image or text, $s$ is the cosine similarity function and N is the batch size and M-1 is the number of resemblant words of one label. As can be seen in Fig. \ref{sitm_cam}, the parallel attention layer focuses on the main character features in the image to guide the matching procedure.

During the inference stage, we only calculate the image-to-text(i2t) similarity to find the highest score among candidates from the candidate set:
\begin{equation}
    p^{i2t}_m(I) = \frac{{\rm exp}(s(I, T_m) / \tau)}{\sum_{n=1}^{T} {\rm exp}(s(I, T_n) / \tau},
\end{equation}
where T is the number of candidates in the candidates set which is depicted in Section 3.1.

\subsection{Overall Objective Function}
A supervised cross-entropy loss function is utilized for training and optimizing the scene text recognizer. By minimizing the negative log-likelihood sequence probability loss function, the difference between the prediction and the ground truth is quantified. The specific formula is given below:
\begin{equation}
    {\mathcal{L}}_{recog} = {\bm E}_{(x, y) \sim (X, Y)} {\left\{- {\rm log}p(y |F_{v}(x) \right\}}.
\end{equation}
The full loss function of the Scene Image-Text Matching module consists of ${\mathcal{L}}_{itc}$ 
\begin{equation}
    {\mathcal{L}}_{SITM} = {\mathcal{L}}_{itc},
\end{equation}
\begin{equation}
    {\mathcal{L}}_{itc} = \frac{1}{2}{\bm E}_{(I, T) \sim D} 
    {\left[{\rm \bm H}({\bm y}^{i2t}(I), {\bm p}^{i2t}(I)) + {\rm \bm H}({\bm y}^{t2i}(T), {\bm p}^{t2i}(T))
    \right]},
\end{equation}
where ${\bm y}^{i2t}(I)$ and ${\bm y}^{t2i}(T)$ denote the ground-truth one-hot similarity, in which the negative pair probability is 0 and the positive pair probability is 1. The H is defined as the cross-entropy loss.



The overall objective function ${\mathcal{L}}_{overall}$ is defined as:
\begin{equation}
    {\mathcal{L}}_{overall} = \lambda_1{\mathcal{L}}_{recog} + \lambda_2{\mathcal{L}}_{SITM},
\end{equation}
where $\lambda_1$ and $\lambda_2$ are the hyper-parameters used to control the training stages. We respectively set $\lambda_1 = 1, \lambda_2 = 0$ and $\lambda_1 = 0 , \lambda_2 = 1$ when we train the recognition module and Scene Image Text Matching module.

\section{Experimental Results}
\subsection{Datasets}
The common synthetic datasets SynthText\cite{ref_lncs49} and MJSynth\cite{ref_lncs50} are utilized to train our proposed model. We employ six widely used benchmarks to evaluate the performance of the model, including three regular text datasets ICDAR2013, SVT, IIIT5K and three irregular text datasets ICDAR2015, SVTP and CUTE80. Following are the specifics of the datasets:

ICDAR2013(IC13)\cite{ref_lncs52} has 1015 test images. The dataset contains only horizontal text instances.

Street View Text (SVT)\cite{ref_lncs53} contains 647 images collected from Google Street View. This dataset contains fuzzy, blurry, and low-resolution text images.

IIIT5K\cite{ref_lncs51} contains 3000 test images crawled from Google image searches with query words. Most text instances are rules for horizontal layout.

ICDAR2015(IC15)\cite{ref_lncs54} contains 1811 test images created for the ICDAR 2015 Robust Reading competitions. Most instances of text are irregular (noisy, blurry, perspective or curved). 

Street View Text Perspective (SVTP)\cite{ref_lncs55} contains 645 cropped images from Google Street View. Many of the images have a distorted perspective.

CUTE80\cite{ref_lncs56} is collected from nature scenes and contains 288 cropped images for verification. Most of them are curved text.

\subsection{Training Setting}
PyTorch is applied to implement the model proposed in this paper. All the experiments are conducted on a 24GB-memory NVIDIA3090. All input images are scaled to $32\times128$ while maintaining their aspect ratio.The character set includes 37 classes, which contains 10 digits, 26 lowercase letters, and an EOS token. The maximum sequence length is 25. Adam is selected as the optimizer, and the batch size is set to 320.

The training procedure consists of two stages: the recognition stage and the matching stage. In the recognition stage, the text recognition network is merely trained to minimize the text recognition loss function. We trained the recognition network 8 epochs on SynthText and MJSynth from scratch. During the matching stage, the SITM network is unfrozen. Image-Text contrastive loss is applied to train the text encoder.
\subsection{Comparison with the Ordinary Dictionary Method and the State-of-the-art}
In this part, we compare the accuracy of the baseline, ordinary dictionary-guided baseline and the proposed dictionary-guided baseline on the six benchmarks. The baseline described in Section 3.1 serves as a comparison standard in our experiments. We utilize the same lexicon, which comprises approximately 20K words and is composed of numbers, common English words and common English trademarks. \textit{Full Lexicon} is not utilized to construct the dictionary, which means that some words in the test set may not be in the lexicon. As we consider this would be a more realistic dictionary composition with some words included and some were excluded. For a fair comparison, all the methods are trained on the SynthText and MJSynth datasets.
\begin{table}[H]
\caption{Comparison with Ordinary Dictionary-guided Baseline.}
\centering
\begin{tabular}{lccclccclc}
\hline
\multicolumn{1}{c}{\multirow{2}{*}{Method}} & \multicolumn{3}{c}{Regular Text}              &  & \multicolumn{3}{c}{Irregular Text}            &  & \multirow{2}{*}{Average} \\ \cline{2-4} \cline{6-8}
\multicolumn{1}{c}{}                        & IC13          & SVT           & IIIT5K        &  & IC15          & SVTP          & CUTE80        &  &                          \\ \hline
Baseline                                    & 94.9          & 90.4          & 94.6          &  & 81.7          & 84.2          & 86.5          &  & 89.8                     \\
Baseline+Dict Guided                        & 95.8          & 92.1          & 96.2          &  & 85.6          & 87.4          & 90.8          &  & 92.1                     \\
Baseline+Our Method                         & \textbf{97.8} & \textbf{94.1} & \textbf{97.1} &  & \textbf{88.0} & \textbf{89.3} & \textbf{93.4} &  & \textbf{93.8}            \\ \hline
Improvement                                 & \textbf{+2.0} & \textbf{+2.0} & \textbf{+0.9} &  & \textbf{+2.4} & \textbf{+1.9} & \textbf{+2.6} &  & \textbf{+1.7}            \\ \hline
\end{tabular}
\label{table_comparison_baseline}
\end{table}

As can be seen from Table \ref{table_comparison_baseline}, utilizing ordinary dictionary guidance would enhance performance, but the improvement on some benchmarks is insignificant. On six benchmarks, our proposed dictionary-guided method outperforms the ordinary method with 2.0\%, 2.0\%, 0.9\%, 2.4\%, 1.9\% and 2.6\% on IC13, SVT, IIIT5K, IC15, SVTP and CUTE80 datasets, respectively. We also discover that our method has superiority on irregular datasets IC15, SVTP and CUTE80 as they contain low-quality images such as curved and blurred texts. As the visual prediction of the irregular datasets often have more severe deviation from the ground truth, the candidate with smallest edit distance may not be the correct answer. 

\begin{figure}[]
\centering
\includegraphics[width=1\linewidth]{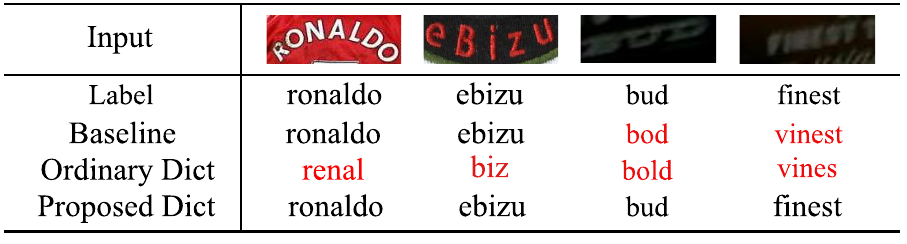}
\caption{Qualitative results for the ordinary dictionary method and our proposed method.}
\label{result_compartision}
\end{figure}

The weaker performance of the ordinary dictionary method stems from two aspects. 1) It disregards visual prediction. The ordinary dictionary pipeline takes the word in the dictionary as output with the smallest edit distance for the visual prediction that is not a component of the dictionary, which makes the correct prediction incorrect. 2) For words in the dictionary with the same edit distance, the traditional dictionary pipeline is unable to determine which output is right. The random selection will fail to output the correct outcome among the candidates.\\ 
\indent Our proposed pipeline effectively avoids the aforementioned issues. In addition to the text recognition network, we also train a SITM network. When a dictionary is employed, we combine the prediction and the top N smallest edit distance dictionary words as candidates set, and the SITM network is used to determine which one is correct. Fig. \ref{result_compartision} illustrates instances successfully recognized by our method while ordinary dictionary method could not. The second and third columns represent that the visual prediction of the scene text recognition network is correct, but there is no corresponding in the dictionary. In this case, the ordinary method generates the wrong answer. However, our proposed method can find the visual prediction output in candidates. The fourth and fifth columns represents the deviation between the visual predictions and the ground truths. When facing candidates with the same edit distance, the ordinary method can only randomly output, while our proposed method can find the correct candidate word according to the SITM network.
\begin{table}[H]
\caption{Comparison with State-of-the-art Methods and Ordinary Dictionary-guided State-of-the-art Methods.}
\begin{tabular}{cccccccccc}
\hline
\multirow{2}{*}{Methods} & \multirow{2}{*}{\begin{tabular}[c]{@{}c@{}}Ordinary \\ Dict Guide\end{tabular}} & \multirow{2}{*}{\begin{tabular}[c]{@{}c@{}}Proposed \\Dict Guide\end{tabular}} & \multicolumn{3}{c}{Regular Text}              &  & \multicolumn{3}{c}{Irregular Text}            \\ \cline{4-6} \cline{8-10} 
                         &                                                                               &                                                                              & IC13          & SVT           & IIIT5K        &  & IC15          & SVTP          & CUTE80        \\ \hline
PlugNet\cite{mou2020plugnet}                  & -                                                                             & -                                                                            & 95.0          & 92.3          & 94.4          &  & 82.2          & 84.3          & 85.0          \\
SRN\cite{ref_lncs33}                      & -                                                                             & -                                                                            & 95.5          & 91.5          & 94.8          &  & 82.7          & 85.1          & 87.8          \\
RobustScanner\cite{yue2020robustscanner}            & -                                                                             & -                                                                            & 94.1          & 89.3          & 95.4          &  & 79.2          & 82.9          & 92.4          \\
TextScanner\cite{wan2020textscanner}              & -                                                                             & -                                                                            & 94.9          & 92.7          & 95.7          &  & 83.5          & 84.8          & 91.6          \\
AutoSTR\cite{zhang2020autostr}                  & -                                                                             & -                                                                            & 94.2          & 90.9          & 94.7          &  & 81.8          & 81.7          & -             \\
VisionLAN\cite{wang2021visionlan}                & -                                                                             & -                                                                            & 95.7          & 91.7          & 95.8          &  & 83.7          & 86.0          & 88.5          \\
CRNN\cite{baek2019wrong}                     & -                                                                             & -                                                                            & 88.8          & 78.9          & 84.3          &  & 61.5          & 64.8          & 61.3          \\
ABINet\cite{ref_lncs44}                   & -                                                                             & -                                                                            & \textbf{97.4}          & 93.5          & 96.2          &  & 86.0          & \textbf{89.3}          & 89.2          \\
PARSeq\cite{ref_lncs18}                   & -                                                                             & -                                                                            & 97.0          & \textbf{93.6}          & \textbf{97.0}          &  & \textbf{86.5}          & 88.9          & \textbf{92.2}          \\ \hline
CRNN\cite{baek2019wrong}                     & $\checkmark$                                                                  &                                                                              & 95.2          & 90.8          & 91.5          &  & 83.0          & 84.0          & 78.5          \\
CRNN\cite{baek2019wrong}                     & -                                                                             & $\checkmark$                                                                 & \textbf{96.9} & \textbf{92.1} & \textbf{93.2} &  & \textbf{84.6} & \textbf{88.5} & \textbf{80.9} \\
ABINet\cite{ref_lncs44}                   & $\checkmark$                                                                  &                                                                              & 97.7          & 94.1          & 96.8          &  & 87.5          & 90.0          & 90.3          \\
ABINet\cite{ref_lncs44}                   & -                                                                             & $\checkmark$                                                                 & \textbf{98.4} & \textbf{95.8} & \textbf{98.0} &  & \textbf{88.6} & \textbf{90.1} & \textbf{91.3} \\ \hline
\end{tabular}
\label{table_comparison_sota}
\end{table}

To verify the effectiveness of our method, we combine two existing scene text recognition frameworks with our proposed dictionary method. We select the CRNN and the state-of-the-art ABINet to validate our proposed approach.\\
\indent Table \ref{table_comparison_sota} shows that our proposed method still outperforms the ordinary dictionary method. As can be seen from the comparison, in the CRNN\cite{baek2019wrong}, our proposed dictionary-guided method outperforms the ordinary method with 1.7\%, 1.3\%, 1.7\%, 1.6\%, 4.5\% and 2.4\% on IC13, SVT, IIIT5K, IC15, SVTP, CUTE80 datasets, respectively. In ABINet\cite{ref_lncs44}, the improvements on the six benchmarks are 0.7\%, 1.7\%, 1.2\%, 1.1\%, 0.1\% and 1.0\%, respectively. In the meanwhile, we find that the utilization of a dictionary to rectify the visual prediction is a highly effective way of enhancing performance. When employing a dictionary to rectify visual prediction, the CRNN\cite{baek2019wrong} exceeds numerous state-of-the-art methods on some benchmarks.

\subsection{Ablation Study}

\begin{table}[H]
\caption{Comparison of recognition accuracy on different numbers of candidates.}
\centering
\begin{tabular}{c|ccccccccccccclc}
\hline
candidates                                                      & 1    &  & 5    &  & 10   &  & 20   &  & 30   &  & 80   &  & 150  &  & 300  \\ \hline
\begin{tabular}[c]{@{}c@{}}Recognition \\ Accuracy\end{tabular} & 92.1 &  & 93.8 &  & 94.1 &  & 94.2 &  & 94.2 &  & 94.3 &  & 94.3 &  & 94.3 \\ \hline
\end{tabular}
\label{table_comparison_candidates}
\end{table}

\subsubsection{The recognition accuracy of baseline as the numbers of candidates varies:}
The quantity of candidates is one of the primary distinctions between our approach and the ordinary pipeline. Table \ref{table_comparison_candidates} demonstrates how the amount of candidate words affects the accuracy of the pipeline. The second column, when the number of candidates is 1, corresponds to the ordinary dictionary-guided method. As can be observed, a substantial improvement of 2\% in accuracy occurs when the candidate number increases from 1 to 10, which explains that the correct word is not necessarily the one with the smallest edit distance. The average accuracy marginally improves as the number of candidates increases from 10 to 80. The saturation appears when the number arrives at 150. Table 3 illustrates the primary benefit of the proposed method, which can select the correct output from a group of options.

\begin{table}[H]
\centering
\caption{Comparison of recognition accuracy on different numbers of resemblant words.}
\begin{tabular}{c|clclclclc}
\hline
\begin{tabular}[c]{@{}c@{}}Number of \\ resemblant words\end{tabular} & 0    &  & 3    &  & 7    &  & 15   &  & 31   \\ \hline
\begin{tabular}[c]{@{}c@{}}Recognition \\ Accuracy\end{tabular}       & 91.1 &  & 93.8 &  & 93.9 &  & 93.9 &  & 93.9 \\ \hline
\end{tabular}
\label{table_comparison_resemblant_words}
\end{table}

\subsubsection{The discussion of resemblant word function:}
For image-text pairs to be successfully matched, a certain number of hard negatives are included in the training process. To illustrate the efficacy of this strategy, we arrange a variety of resemblant words: 0, 3, 7, 15 and 31. The recognition accuracy of the entire pipeline is shown in Table \ref{table_comparison_resemblant_words}. The second column 0 indicates that no hard negative is used. It can be seen that recognition accuracy improves as the number of hard negatives increases. However, it will not be improved until a certain number of hard negatives has been accumulated. In contrast, when the number of hard negatives is equal to 0, the SITM network cannot complete the image-text matching task, therefore some incorrect matching pairs are produced. The performance(91.1\% accuracy) is significantly worse than the ordinary dictionary method performance(92.1\% accuracy).  Table 4 demonstrates that the model is capable of learning more fine-grained distinctions between different text features through resemblant word generation strategy.

\section{Conclusion}
In this paper, we propose a new dictionary-guided scene text recognition method, which integrates the visual features into the inference stage and can effectively boost the performance of dictionary language model. In addition, the SITM is designed to indicate the correctness of explicit language model rectification. The resemblant words generation strategy, which utilizes labels to generate hard negatives in the training stage, is presented to improve the matching accuracy of SITM network. The experiments on six mainstream benchmarks demonstrate that our method outperforms the ordinary dictionary method and also show superiority in other state-of-the-art scene text recognition methods.

\section*{Acknowledgement}
This work is supported by  the National Natural Science Foundation of China under Grant No. 62176091.

%
%
%
\bibliographystyle{splncs04}
\bibliography{ref}

\end{document}